\documentclass[10pt,twocolumn,letterpaper]{article}

\usepackage[pagenumbers]{cvpr} 

\usepackage{graphicx}
\usepackage{amsmath}
\usepackage{amssymb}
\usepackage{booktabs}

\usepackage{algorithm}
\usepackage{algpseudocode}

\usepackage[pagebackref,breaklinks,colorlinks]{hyperref}

\usepackage[capitalize]{cleveref}
\crefname{section}{Sec.}{Secs.}
\Crefname{section}{Section}{Sections}
\Crefname{table}{Table}{Tables}
\crefname{table}{Tab.}{Tabs.}

\newcommand{\fref}[1]{Figure~\ref{#1}}
\newcommand{\tref}[1]{Table~\ref{#1}}

\def\cvprPaperID{*****} 
\def\confName{CVPR}
\def\confYear{2023}

\begin{document}

\title{Random Position Adversarial Patch for Vision Transformers}

\author{Mingzhen Shao\\
Kahlert School of Computing\\
University of Utah\\
{\tt\small shao@cs.utah.edu}}

\maketitle

\begin{abstract}

 
 Previous studies have shown the vulnerability of vision transformers to adversarial patches, but these studies all rely on a critical assumption: the attack patches must be perfectly aligned with the patches used for linear projection in vision transformers. 
 Due to this stringent requirement, deploying adversarial patches for vision transformers in the physical world becomes impractical, unlike their effectiveness on CNNs. 
 This paper proposes a novel method for generating an adversarial patch (G-Patch) that overcomes the alignment constraint, allowing the patch to launch a targeted attack at any position within the field of view. 
 Specifically, instead of directly optimizing the patch using gradients, we employ a GAN-like structure to generate the adversarial patch.
 Our experiments show the effectiveness of the adversarial patch in achieving universal attacks on vision transformers, both in digital and physical-world scenarios. Additionally, further analysis reveals that the generated adversarial patch exhibits robustness to brightness restriction, color transfer, and random noise. 
 Real-world attack experiments validate the effectiveness of the G-Patch to launch robust attacks even under some very challenging conditions. 
 



\end{abstract}

\section{Introduction}
\label{sec:intro}
 Recently, vision transformers (ViTs) have garnered significant attention due to their impressive performance~\cite{dosovitskiy2020image, chen2021crossvit, chen2021visformer, graham2021levit, han2021transformer, liu2021swin, touvron2021training, xiao2021early} and their ability to surpass convolutional neural networks (CNNs) in various domains. 
 This remarkable performance has spurred interest in examining the robustness of ViTs, particularly considering the well-known vulnerability of CNNs to adversarial attacks~\cite{bhojanapalli2021understanding, qin2022understanding, salman2022certified, shi2022decision}. 

 Drawing from the lessons learned with CNNs, 
 adversarial attacks can be classified as image-dependent or image-independent.
 An image-dependent attack typically makes minute modifications to the source image~\cite{goodfellow2014explaining, madry2017towards, wu2020skip}. 
 These approaches have the drawback of being tailored to specific source images and limited in their deployment in the physical domain. 
 Moreover, vision transformers have demonstrated remarkable robustness against these types of attacks\cite{bhojanapalli2021understanding}, showing their resilience when facing such adversarial attacks.
 

 In contrast, the image-independent attack aims to create a patch that can be put alongside the target, without prior knowledge of lighting conditions, camera angles, or even elements present in the scene~\cite{brown2017adversarial}. 
 Adversarial patches have proven highly effective against CNNs, as they can be positioned anywhere within the classifier's field of view to launch an attack. Astonishingly, they only require 10\% of the pixels in the input image to deceive powerful CNN models like ResNet50~\cite{brown2017adversarial}.

 
 Unlike CNNs, vision transformers treat the input image as a sequence of image patches. 
 To carry out an adversarial patch attack, a commonly employed approach is to substitute certain input image patches with adversarial samples~\cite{fu2022patch, gu2022vision, joshifew}. These studies have shown the heightened vulnerability of vision transformers to adversarial patches. However, all the experiments conducted so far have been limited to the digital domain to accurately locate the adversarial patches. Gu~\etal demonstrated that even a slight shift of a single pixel could dramatically decrease the attack success rate. 
 
 
 To overcome these strict limitations and enable physical-world deployment, we propose a novel approach that uses a GAN-like structure to generate universal and targeted adversarial patches (G-Patch). 
 Our model consists of three main components: the generator, deployer, and discriminator. 
 The generator is responsible for creating an adversarial patch. 
 The deployer then attaches the patch to a random position on the source image. 
 Finally, the victim network acts as the discriminator, providing predictions based on the modified image. 
 Notably, unlike traditional GAN setups, the discriminator (victim network) remains unaltered throughout the training process. 
 
 Our experiments demonstrate that the generated adversarial patches can successfully launch attacks on various victim models at any position within the field of view. These patches achieve a high attack success rate of over 90\% while maintaining a small size of approximately 10\% of the source image (on ViT-B/16).
 Further analysis reveals that the generated adversarial patches exhibit strong robustness to brightness restriction, color transfer, and random noise. This robustness to different distributions enhances their effectiveness during physical-world deployment.
 To assess their practical performance, we printed and positioned the adversarial patches in real-world settings. The results show that the patches consistently perform robustly in the physical world.
 To the best of our knowledge, our research is the first endeavor to achieve random position adversarial patch attacks on vision transformers.
 

 Our contributions can be summarized as follows:
 \begin{itemize}
     \item We propose a new model to generate random position adversarial patches for vision transformers, which can launch targeted attacks at any position within the field of view. 
     \item We show that the adversarial patches generated for vision transformers exhibit strong robustness to brightness restriction, color transfer, and random noise.
     \item We demonstrate that the generated adversarial patch can be robustly deployed in the physical world.
 \end{itemize}
 
\begin{figure*}[t]
    \centering
    \includegraphics[width=0.9\linewidth]{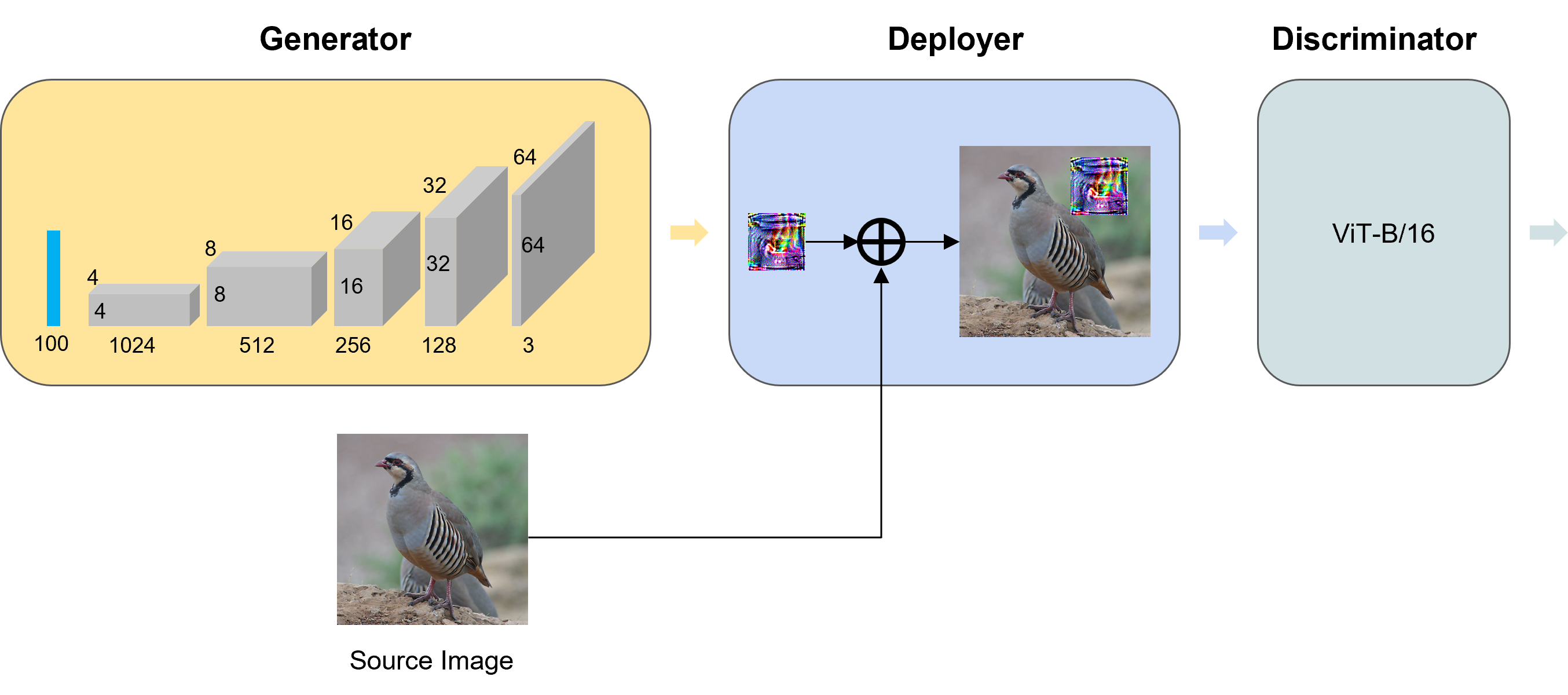}
     \caption{Network}
     \label{fig:network}
\end{figure*}

\section{Related work}
\label{sec:related_work}
 \subsection{Vision transformer}
 The transformer was first introduced by Vaswani~\etal~\cite{vaswani2017attention} for NLP tasks. 
 Following the success in NLP, Dosoviskiy~\etal~\cite{dosovitskiy2020image} proposed the vision transformer (ViT) that leveraged non-overlapping patches as tokens input to a similar attention based architecture. 
 Since then, numerous models have been proposed to improve the performance of vision transformer models. 
 Touvron~\etal~\cite{touvron2021training} introduced a teacher-student strategy in their DeiT models that dramatically reduced the pre-training request. 
 Liu~\etal~\cite{liu2021swin} proposed the SWIN transformer using the shifted windowing scheme that achieves greater efficiency
 by limiting self-attention computation to non-overlapping local windows while also allowing for cross-window connection. 

 Vision transformers have also been used in different vision tasks, including zero-shot classification~\cite{radford2021learning}, captioning~\cite{li2022blip}, and image generation~\cite{rombach2022high}. 
 Due to the great success of vision transformers on version tasks, many researchers have committed to analyzing the models' robustness to adversarial attacks.

 \subsection{Adversarial attacks}
 Adversarial attacks are widely employed to deceive deep learning models, resulting in remarkable successes.
 The first adversarial attack for deep learning was introduced by szegedy~\etal~\cite{szegedy2013intriguing}. 

 Since their seminal work, numerous researchers have devised increasingly efficient techniques for generating adversarial attacks. 
 In computer vision tasks, adversarial attacks can be classified into two types, depending on their reliance on the input image.
 
 The first type is image-dependent adversarial attacks, which typically make minute modifications to the source image. 
 These attacks employ various optimization strategies such as 
 the Fast Gradient Sign Method (FGSM)~\cite{goodfellow2014explaining}, Projected Gradient Descent (PGD)~\cite{madry2017towards}, and Skip Gradient (SGD)~\cite{wu2020skip}. 
 However, these approaches often exhibit weaknesses due to their design being tailored to specific source images or limited to the digital domain. 

 In contrast, the second type of attack, image-independent attacks, uses an additional object (patch) to eliminate the requirement of relying on the input image.
 The patch is trained to create an attack without prior knowledge of the other items within the scene. 
 The first image-independent attack approach was proposed by Brown~\etal~\cite{brown2017adversarial}. They used gradient-based optimization to iteratively update the pixel values of the patch in the source image to find the optimal values that can cause the target model to misclassify the object (AdvPatch). 
 This patch can be placed anywhere within the field of view of a classifier and launch an attack. 
 Since then, many studies have followed the same strategy to develop patches for physical attacks aimed at deceiving classifiers or object detectors, such as traffic signs~\cite{evtimov2017robust}, cloaks~\cite{wu2020making}, or vehicles~\cite{zhang2019camou}. 
 These successes have sparked researchers' interest in applying the same methods to vision transformers.

 \subsection{Robustness of vision transformer}
 Shortly after the introduction of vision transformers, several researchers~\cite{bhojanapalli2021understanding, paul2022vision, shao2021adversarial} conducted studies demonstrating the superior robustness of vision transformers compared to CNNs when the entire image is perturbed with adversarial perturbations. 
 However, subsequent research by Fu~\etal~\cite{fu2022patch} explored the vulnerability of vision transformers to patch attacks and found that vision transformers are more susceptible to such attacks compared to CNNs. 
 Additionally, Gu~\etal~\cite{gu2022vision} further showed that whereas vision transformers are generally resilient to patch-based natural attacks, they are more vulnerable to adversarial patch attacks when compared to comparable CNNs.

 All the preceding studies used a generation method similar to the one employed for CNNs,
 which involves replacing certain input patches with random noise and uses gradient-based optimization to iteratively update the pixel values, aiming to find the optimal values that can deceive the target model.
 However, unlike adversarial patches for CNNs, the patches they obtained for vision transformers must be precisely aligned with the input image patches. 
 Gu~\etal demonstrated that even a slight shift of a single pixel could dramatically decrease the attack success rate. 
 The strict requirement posed a significant challenge to the practicality of using adversarial patches in real-world scenarios, as misalignment between attack patches and image patches is commonly encountered due to various factors.
 Consequently, it is crucial to develop methods for creating adversarial patches that account for these realistic conditions where misalignment can occur.

\section{Methods}

\subsection{Network}

 We introduce a GAN-like model for generating the adversarial patch. Instead of relying on direct gradient optimization of a random initial patch, our approach employs a five-layer convolutional network to transform a random noise into the desired adversarial patch. 
 The structure of our model is shown in \fref{fig:network}.
 
 The model can be divided into three functional parts: generator, deployer, and discriminator. 
 
 \textbf{Generator:} 
 The generator consists of five convolutional layers, each accompanied by batch normalization and ReLU activation layers. The first convolutional layer is responsible for projecting and reshaping the random input vector into a four-dimensional tensor. The next four convolutional layers progressively upsample and refine the feature maps, capturing more complex patterns and details.
 The last convolutional layer is followed by a threshold layer instead of the batch normalization and ReLU layers. This threshold layer ensures that the output values of the generator are limited to a specific range. 
 The threshold layer is defined as follows:
 \begin{equation}
     Th(x) = k * tanh(x) + k
 \end{equation}
 where $k$ is a hyperparameter to adjust the range of the output and $tanh(x)$ applies the hyperbolic tangent function element-wise. We add $k$ here to ensure that all values in the output remain above 0.
 
 In default, we use $k=0.5$ to scale the range of the patch to 1, and for the following experiments, we use different $k$ to achieve brightness restriction. 
 By changing the kernel size and stride of different convolutional layers, we can change the size of the output adversarial patch. 

 \textbf{Deployer:} Given an image $x\in {[0,1]}^{w \times h \times c}$ with class $y$ and the generated adversarial patch $p$. We use Algorithm~\ref{alg:mask} to generate a random binary mask $M$ with the same shape of $x$.

\begin{algorithm}[H]
    \caption{Mask generation}\label{alg:mask}
    \hspace*{\algorithmicindent} \textbf{Input} {source image: $x$, adversarial patch: $p$}
    \begin{algorithmic}[1]       
        \State $M \gets \text{zeros}(x)$ \Comment{all-zero mask $M$ with shape $x$}
        \State $k \gets \text{random.randint}(0, M[0]-p[0])$ 
        \State $l \gets \text{random.randint}(0, M[1]-p[1])$ \Comment{random position $k$,$l$ within the range of $M$}
        \For{$i \gets k$ \textbf{to} $k+p[0]-1$}
            \For{$j \gets l$ \textbf{to} $l+p[1]-1$}
                \State $M[i,j] \gets 1$ \Comment{mask only includes elements within shape $p$ at position $(k, l)$}
            \EndFor
        \EndFor
        \State \textbf{return} $M$
    \end{algorithmic}
\end{algorithm}

 Then a modified image is generated by the deploy function $T(p,x)$:
 \begin{equation}
     T(p,x) = M*p + (1-M)*x
 \end{equation}
The modified image is used as input for the discriminator.
 
 \textbf{Discriminator:} The discriminator in our network is composed of the victim network (ViT-B/16 in the figure). It can be replaced with different models to generate adversarial patches for different victim networks. Unlike traditional GANs, the discriminator in our network remains unaltered throughout the training process. 
 
 For a targeted attack, the final loss of our network can be formed as follows:
 \begin{equation}
     L = log(softmax(Pr(\hat{y}|T(p,x))))
 \end{equation}
 where the $\hat{y}$ is the target class and $\hat{y} \neq y$, $Pr$ is the prediction of the discriminator with respect to class $\hat{y}$. 
 

\section{Experimental results and analysis}

 In this section, we first provide detailed information about the experimental setup used in our study. 
 Next, we show the adversarial patches generated by our proposed model and evaluate their performance on various victim networks. Our results highlight the effectiveness of these patches in launching attacks from any position within the field of view.
 Furthermore, we conduct an in-depth analysis of the robustness of the generated adversarial patches. We investigate their resilience to brightness restriction, color transfer, and random noise, providing insights into their stability and effectiveness. 
 Lastly, we validate the practical applicability of the generated adversarial patches by physically printing and placing them in real-world scenarios. This empirical evaluation demonstrates that the patches can effectively deceive vision transformer based systems in complex physical environments.

 \subsection{Experimental setup}
 
 In our experiments, we use the weights and the shared models from the \textit{Pytorch Image models}~\cite{rw2019timm} repository. These models are trained on the ImageNet1K dataset. 
 To ensure the optimal performance, we conduct training for each configuration over 40 epochs, selecting the patch that achieves the highest performance as the final output patch. 
 The input images used in our experiments have dimensions of 224x224, with pixel values from 0 to 1.

 To assess the attack success rate (ASR), we begin by assembling a collection of images that are accurately classified by the models. The total number of these collected images is denoted as $P$.  we apply the adversarial attack patch to this set of images and determine the number of images, denoted as $Q$, that are classified as the target class. The ASR is then defined as $\frac{Q}{P}$, serving as a metric to measure the effectiveness of the attack. 
 
 In order to evaluate the patch's performance in the physical world, we use an HP laser printer to print the adversarial patches on A4 paper. 
 Then we position the printed adversarial patch alongside the target object and capture photographs using a Google Pixel 6a smartphone. 
 This physical-world test incorporates various real-world factors such as camera angle changes, lighting variations, and different types of noise. 
 By subjecting the generated patch to these real-world conditions, we are able to comprehensively evaluate its practical effectiveness in real-world scenarios.




\subsection{Performance of generated patches}
 We select the ViT~\cite{dosovitskiy2020image} and SWIN Transformer~\cite{liu2021swin} as the fundamental victim network structures in our study. These two networks represent the key variations in patch handling within vision transformers: the ViT used fixed, non-overlapping patches, whereas the SWIN transformer incorporates shifted patch sizes. 
 
 The attack success rates of different vision transformers with different patch sizes are summarized in \tref{tab:performance}. 

 \begin{table}[ht]
     \centering
     \begin{tabular}{@{}lccc@{}}
        \toprule
         Models &   \multicolumn{3}{c}{Patch size}  \\ 
                    & 48x48   &   64x64   &   80x80 \\
        \midrule
         ViT-B/16    & 6.7\% & 69.6\%  & 97.1\%  \\
         ViT-L/16   & 2.7\%  & 64.3\%  &  88.7\% \\
         SWIN-B/16   &  59.5\% & 96.8\%  &  99.6\% \\
         \bottomrule
     \end{tabular}
     \caption{ASR of generated patches on different vision transformers}
     \label{tab:performance}
 \end{table}

 We first observe that regardless of the architecture or depth of the vision transformers, the generated adversarial patches can achieve a high attack success rate even with a relatively small size (~10\% of the source image). 
 This finding demonstrates the effectiveness of the generated adversarial patches that can launch attacks from any position of the source image.

 Despite using a distinct method for generating adversarial patches, we are not surprised to discover a strong correlation between the patch size and its performance. Specifically, larger patches exhibit a higher attack success rate. 
 
 When comparing the results on the smaller models (ViT-B/16) to the larger models (ViT-L/16),
 we observe a significant disparity in their robustness across all patch sizes.
 This observation serves as evidence that the larger models possess inherent advantages in terms of defending against such attacks. 

 Furthermore, we confirm that the SWIN-B/16 shows a notably high ASR compared to the same-sized ViT-B/16. 
 This outcome corroborates Shao~\etal's observation~\cite{shao2021adversarial} that emphasizing low-level features in vision transformers can improve their overall performance but may have a detrimental impact on adversarial robustness.
 
\begin{figure}[ht]
    \centering
    \includegraphics[width=0.9\linewidth]{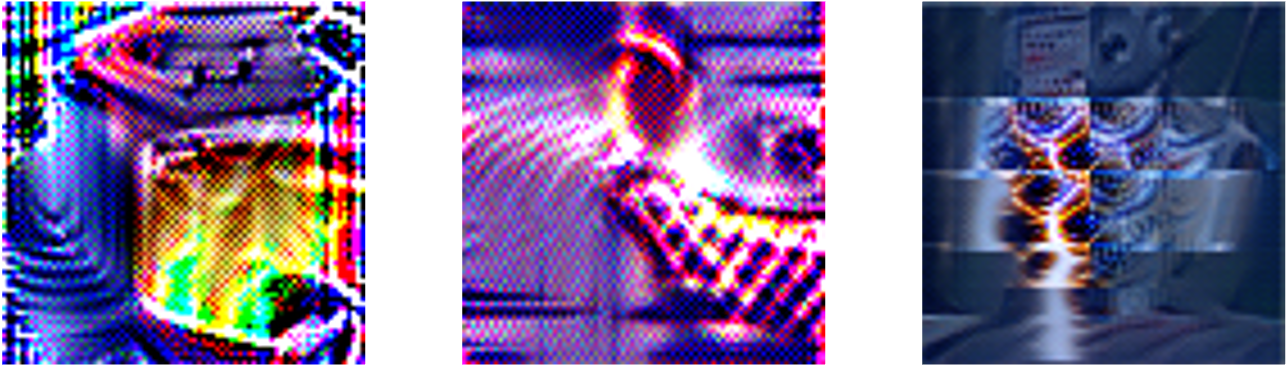}
    \caption{Patches for different networks with size 80x80 {(left: ViT-B/16, middle: ViT-L/16, right: SWIN-B/16)}}
    \label{fig:patch}
\end{figure}

 In Figure \ref{fig:patch}, we present the adversarial patches generated for different victim models. 
 The patch created for the SWIN model displays a remarkable dissimilarity compared to those for ViT models. This striking disparity serves as an explanation for the ineffective transfer of adversarial patches between different models.

\begin{figure}[ht]
     \centering
     \includegraphics[width=0.9\linewidth]{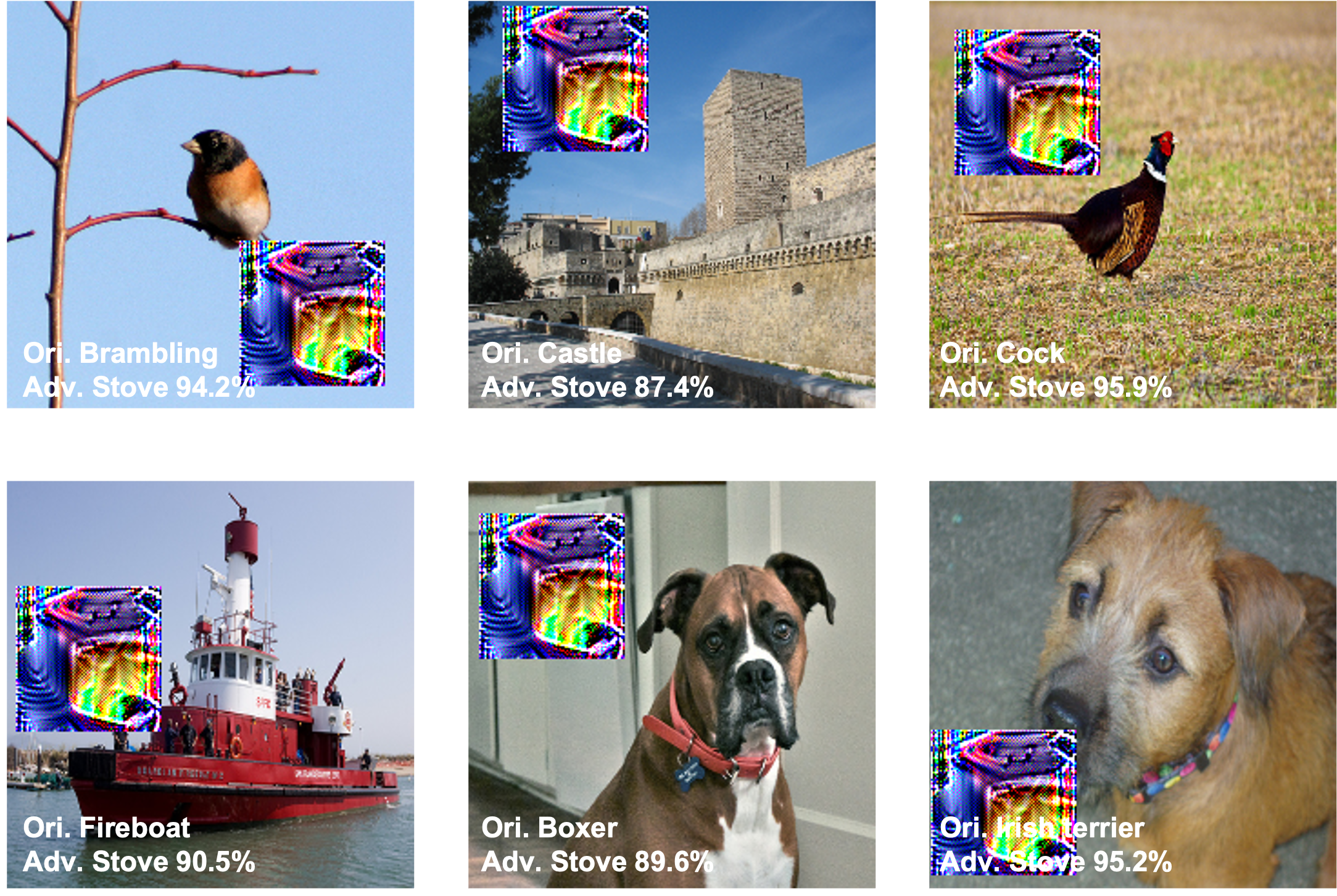}
     \caption{Modified images with random patch position}
     \label{fig:demo}
 \end{figure}

 Some modified images created for the ViT-B/16 are shown in \fref{fig:demo}. 
 These images demonstrate the flexibility of our patch placement methodology, as the adversarial patch is positioned randomly on the source image, regardless of its specific position or alignment. 
 
 \subsection{Patch robustness analysis}
 Recent research~\cite{shao2023brightnessrestricted} has demonstrated the remarkable robustness of adversarial patches designed for CNNs against brightness restriction, color transfer, and random noise.
 Inspired by this finding, we adopt a similar investigative approach to analyze the behavior of patches designed for vision transformers.

 For our experiments, 
 we use the patch size 80x80 to assess the impact of different features on the victim networks (ViT-B/16 and SWIN-B/16).
 
 \subsubsection{Brightness restriction}

\begin{figure}
     \centering
     \includegraphics[width=0.9\linewidth]{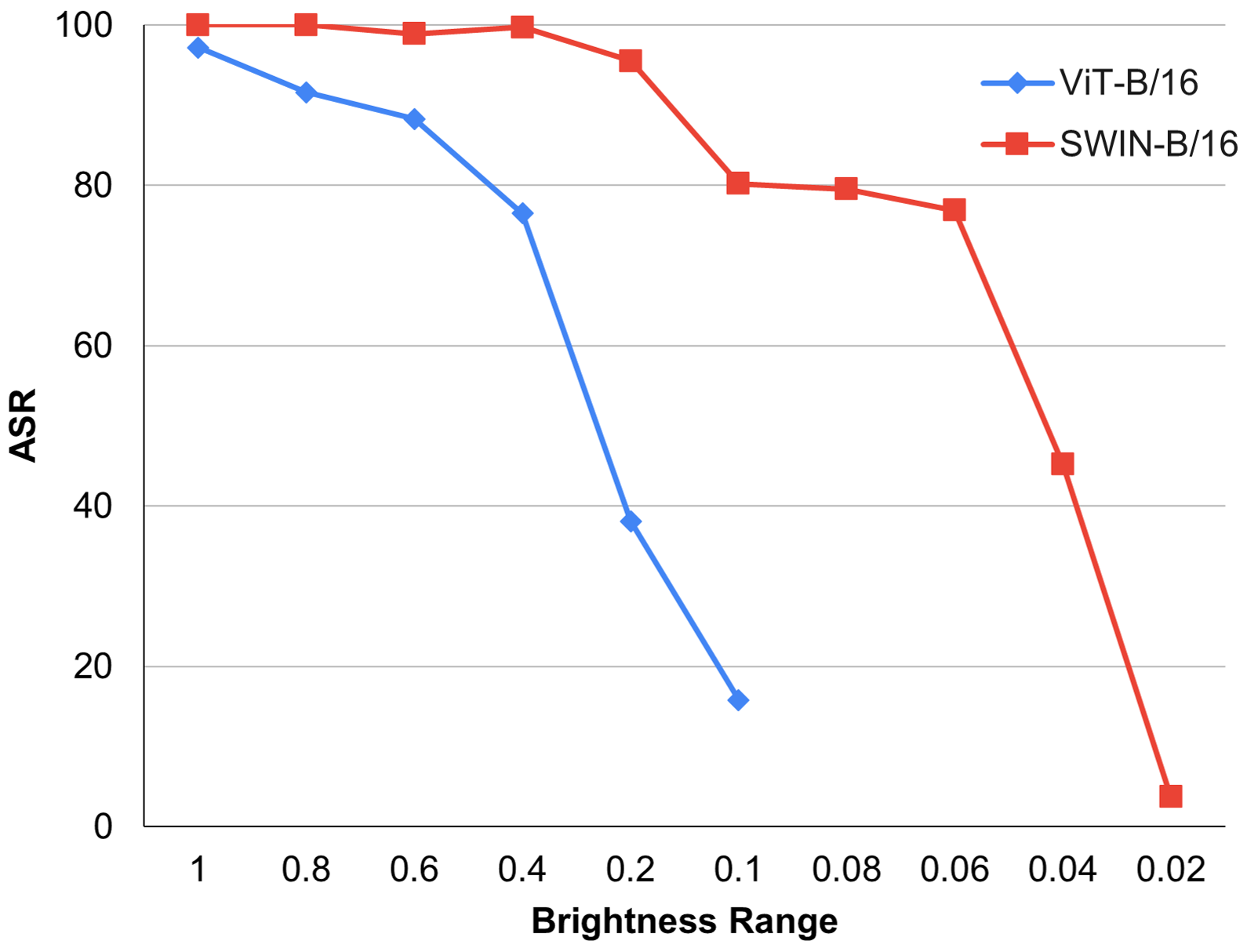}
     \caption{ASR with different brightness range}
     \label{fig:range}
 \end{figure}

\begin{figure*}[ht]
     \centering
     \includegraphics[width=0.9\linewidth]{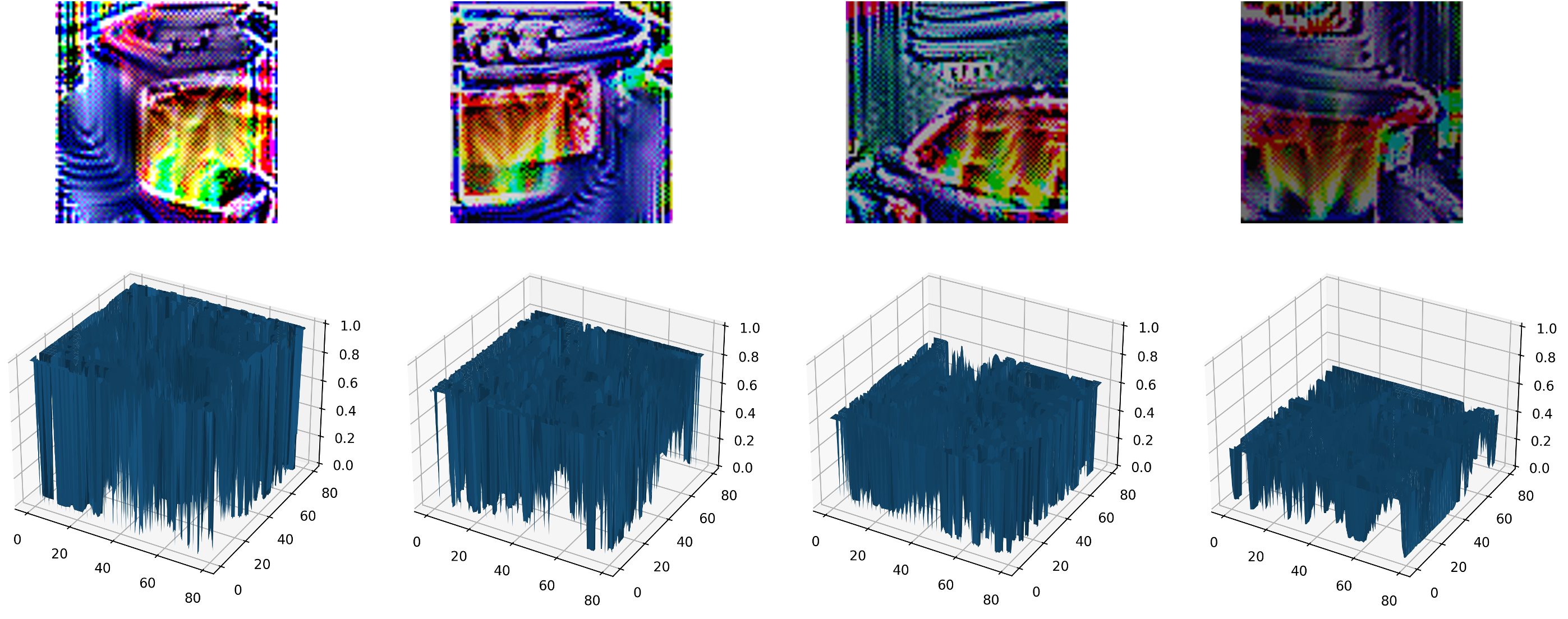}
     \caption{Patches with different brightness restriction and their brightness distribution}
     \label{fig:3D}
 \end{figure*}

 Brightness restriction for the generated patches can be easily implemented by adjusting the $k$ value within the threshold layer. 
 The performance variations for different value ranges are shown in \fref{fig:range}
 We observe a significantly higher robustness in the SWIN model compared to the ViT model when the brightness range is decreased. 
 This disparity in robustness can be attributed to the vulnerability of the SWIN model, as it requires comparatively less information to deceive the network. 
 We find that even when the brightness range diminishes to just half of its original magnitude, the generated patch can still preserve over 80\% of its ASR on ViT-B/16 model. 
 The results show that patches designed for vision transformers exhibit a notable robustness to brightness restriction, similar to those observed in CNNs. 
 Furthermore, the robustness of these patches is closely linked to the structure of the victim network. 
 Some brightness-restricted patches and their brightness distributions are shown in \fref{fig:3D}.

 \subsubsection{Color transfer}
 
  \begin{figure}[ht]
     \centering
     \includegraphics[width=0.9\linewidth]{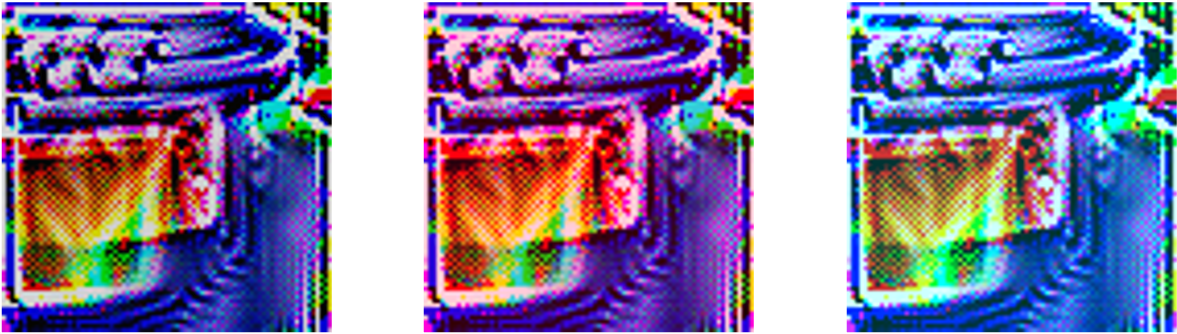}
     \caption{Different color transferred patch on ViT-B/16}
     \label{fig:color}
 \end{figure}
 In order to achieve color transfer, we add a $\delta$ to all values within certain channels. However, performing this operation directly on the original patch (ranging from 0 to 1) can easily result in an overflow.

 To address this issue, we use patches with brightness restriction (ranging from 0 to 0.8). 
 By employing these patches, we ensure that overflow issues were avoided. 
 This color transfer does not alter the texture distribution of the patch.

 We present some color-transferred patches in \fref{fig:color}, and the performance between different color patches is shown in \tref{tab:acc_color}.


 \begin{table}[h]
     \centering
      \begin{tabular}{@{}lccc@{}}
      \toprule
        Models &   \multicolumn{3}{c}{Color}  \\ 
                        & original color   &   color 1  &  color 2 \\
        \midrule
         ViT-B/16        & 91.6\%    & 90.9\%        & 90.7\%  \\
         SWIN-B/16       & 99.6\%    & 98.9\%        & 99.7\%  \\
         \bottomrule
     \end{tabular}
     \caption{ASR with different color transfer}
     \label{tab:acc_color}
 \end{table}

 We find that irrespective of the victim network, the generated adversarial patch consistently achieves nearly identical ASR when subjected to different color transfers. 
 This performance shows that the adversarial patch designed for version transformers does not rely on color information to deceive victim networks.
 This characteristic endows the patch with the capability to diminish its visual appeal through color transfer during deployment, akin to the approach employed by Shao\cite{shao2023brightnessrestricted} for CNNs.

 \subsubsection{Random noise}
 In real-world deployments, the printed patch cannot be precisely the same as the digital version for various reasons (color accuracy of printer, carrier texture). 
 To simulate the noise commonly encountered during real-world deployment, we generate random noise based on different signal-to-noise ratios (SNR).
 In order to avoid overflow when adding strong noise, we choose patches with a narrow brightness range (ranging from 0 to 0.6). 
 The results of our experiments are shown in \tref{tab:noise}. 

  \begin{table}[h]
     \centering
     \begin{tabular}{@{}lcccc@{}}
     \toprule
        Models &   \multicolumn{4}{c}{SNR}  \\ 
                       &    10 dB  &   7 dB   &   5.2 dB   &   4 dB \\
       \midrule
         ViT-B/16   &    83.4\% & 81.6\%  & 69.1\%     & 56.2\%   \\
         SWIN-B/16  &    83.4\% & 81.6\%  & 69.1\%     & 56.2\%   \\
    \bottomrule
     \end{tabular}
     \caption{ASR across different noise levels}
     \label{tab:noise}
 \end{table}

 We observe that the performance of the generated adversarial patch remains relatively stable even at an SNR of 7 dB (20\% random color drift). This finding suggests that the generated adversarial patch exhibits robustness in dealing with the random noise typically encountered in real-world scenarios. The patch's ability to maintain a high attack success rate in the presence of such noise further reinforces its effectiveness and practicality.

 \subsection{Real-world attacks}

 \begin{figure*}[ht]
     \centering
     \includegraphics[width=0.9\linewidth]{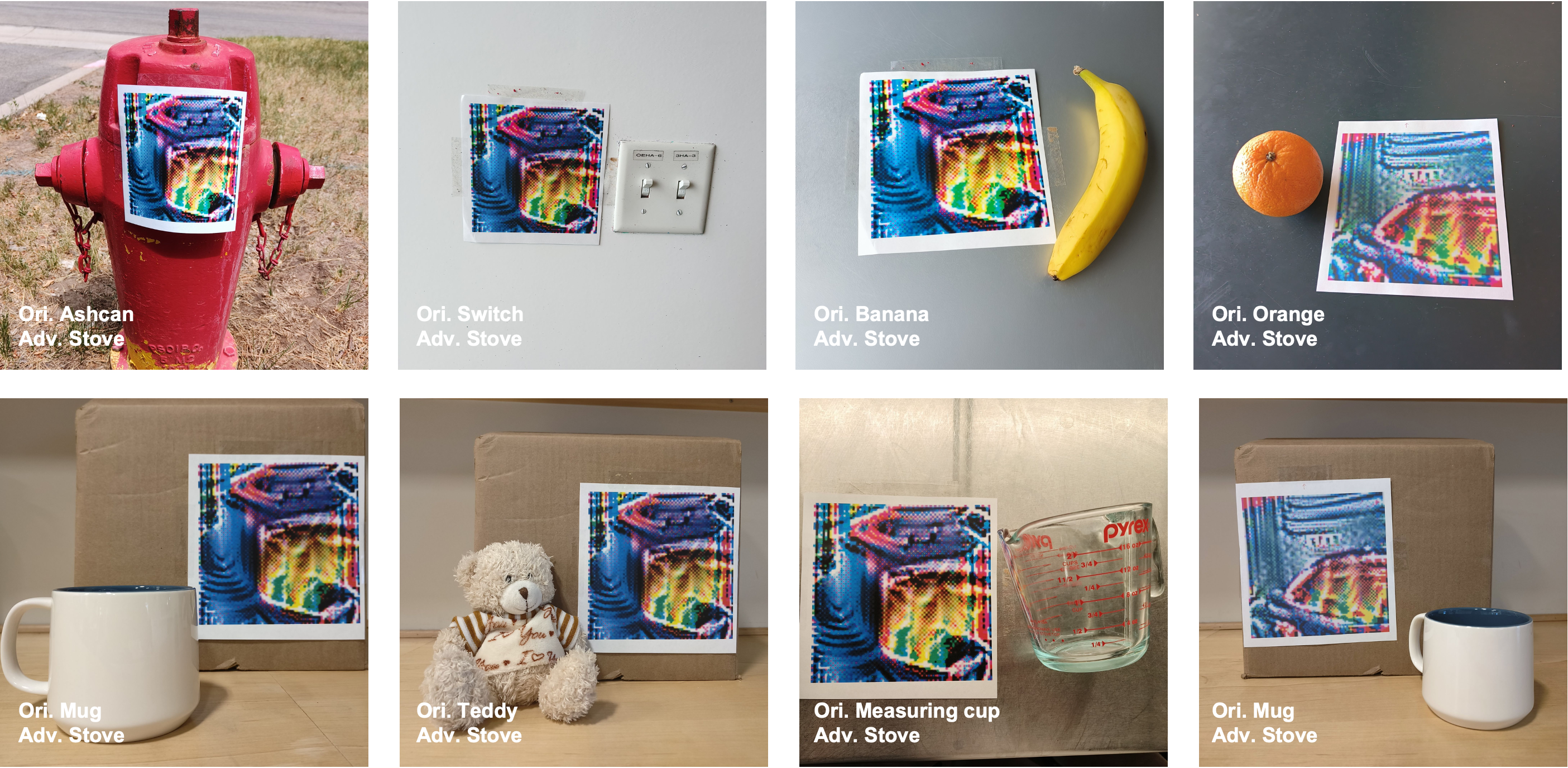}
     \caption{Prediction results in the physical world(top: outdoor lighting, bottom: indoor lighting)}
     \label{fig:physical}
 \end{figure*}
 
 The real-world deployability of adversarial patches designed for CNNs has been demonstrated in many works. However, due to the alignment problem, none of the adversarial patches designed for vision transformers has been deployed in the physical world before. 
 Although our generated adversarial patches show the perfect position irrelative based on previous experiments, a valid concern remains regarding their robustness in real-world scenarios.
 In order to address this concern, we design several real-world deploy instances to show that the proposed attack patch can still work robustly in the physical world. 

 We have selected a range of scenarios, including indoor and outdoor environments, capturing images from different distances, angles, and lighting conditions.
 To ensure a more comprehensive evaluation, we use the ViT-B/16 model as the victim network instead of the easier SWIN-B/16 in our experiments.
 \fref{fig:physical} shows some figures and predictions. 
 
 We find that the generated patches show robust results in the physical world, even with brightness restriction.  
 The top line in \fref{fig:physical} demonstrates the effectiveness of the generated adversarial patches in handling distortions caused by the inclination of the camera angle and even the bending of the printed patch.
 These results show that our generated patch can robustly launch attacks in the physical world without considering the position, lighting, and items in the field of view.

\section{Conclusion}
 This paper introduces the G-Patch generating model, a novel approach for generating random position adversarial patches for vision transformers. 
 The G-Patch successfully attacks on vision transformers from any position within the field of view, without requiring precise alignment or specific position.
 Furthermore, comprehensive analysis reveals that the G-Patch exhibits strong robustness against brightness restriction, color transfer, and random noise. 
 These properties make the G-Patch highly resilient to various disturbances encountered in real-world scenarios.
 Real-world attack experiments validate the effectiveness of the G-Patch, showing its ability to launch robust attacks even under challenging conditions such as large camera angle inclinations and bending of printed patches.
 
 

{\small
\bibliographystyle{ieee_fullname}
\bibliography{egbib}
}

\end{document}